\pdfoutput=1
\documentclass[11pt]{article}

\usepackage{acl}
\usepackage{times}
\usepackage{latexsym}

\usepackage[T1]{fontenc}
\usepackage[utf8]{inputenc}

\usepackage{microtype}
\usepackage{xcolor,colortbl}
\usepackage{multirow}
\usepackage{amsmath}
\usepackage{graphicx}
\usepackage{booktabs}
\usepackage[skins,breakable]{tcolorbox}
\usepackage{color, colortbl, listings}
\usepackage{inconsolata}
\usepackage{enumitem}
\usepackage[capitalize]{cleveref}
\usepackage{graphicx}
\usepackage{subfigure}
\crefname{section}{Sec.}{Secs.}

\newcommand{\method}[0]{\textsc{Soft-SC}}

\newcommand{\fullform}[0]{Soft Self-Consistency}
\definecolor{prmpt}{gray}{0.9}

\DeclareMathOperator*{\argmax}{arg\,max}

\newcommand{\interval}[1]{_{\pm\text{ #1}}}

\title{Soft Self-Consistency Improves Language Model Agents}
\author{Han Wang$^*$ \;\;\;\;\; Archiki Prasad$^*$ \;\;\;\;\; Elias Stengel-Eskin$^*$ \;\;\;\;\;   Mohit Bansal \\
        \textnormal{UNC Chapel Hill} \\ \texttt{\{hwang, archiki, esteng, mbansal\}@cs.unc.edu} \\ }

\begin{document}
\maketitle
\def\thefootnote{*}\footnotetext{Equal Contribution}\def\thefootnote{\arabic{footnote}}
\begin{abstract}
Generations from large language models (LLMs) can be improved by sampling and scoring multiple solutions to select a final answer.
Current ``sample and select'' methods such as self-consistency \citep[SC;][]{wang2023self} rely on majority voting to score answers.
However, when tasks have many distinct and valid answers, selection by voting requires a large number of samples. This makes SC prohibitively expensive for interactive tasks that involve generating multiple actions (answers) sequentially. 
After establishing that majority voting fails to provide consistent gains on such tasks, we demonstrate how to increase success rates by softening the scoring criterion.
We introduce \emph{Soft Self-Consistency} (\method{}), which replaces SC's discontinuous scoring with a continuous score computed from model likelihoods, allowing for selection even when actions are sparsely distributed.
\method{} improves both performance \emph{and} efficiency on long-horizon interactive tasks, requiring half as many samples as SC for comparable or better performance. 
For a fixed number of samples, \method{} leads to a $1.3\%$ increase over SC in absolute success rate on writing bash programs, a $6.6\%$  increase on online shopping (WebShop), and a $4.7\%$ increase for an interactive household game (ALFWorld).
Finally, we show that \method{} can be applied to both open-source and black-box models.\footnote{Our code is publicly available at: \url{https://github.com/HanNight/soft_self_consistency}.}
\end{abstract}

\section{Introduction}
\label{sec:intro}

\begin{figure*}[t]
    \centering
    \includegraphics[trim={0.12cm, 0.38cm, 0.2cm, 0.6cm},clip,width=\linewidth]{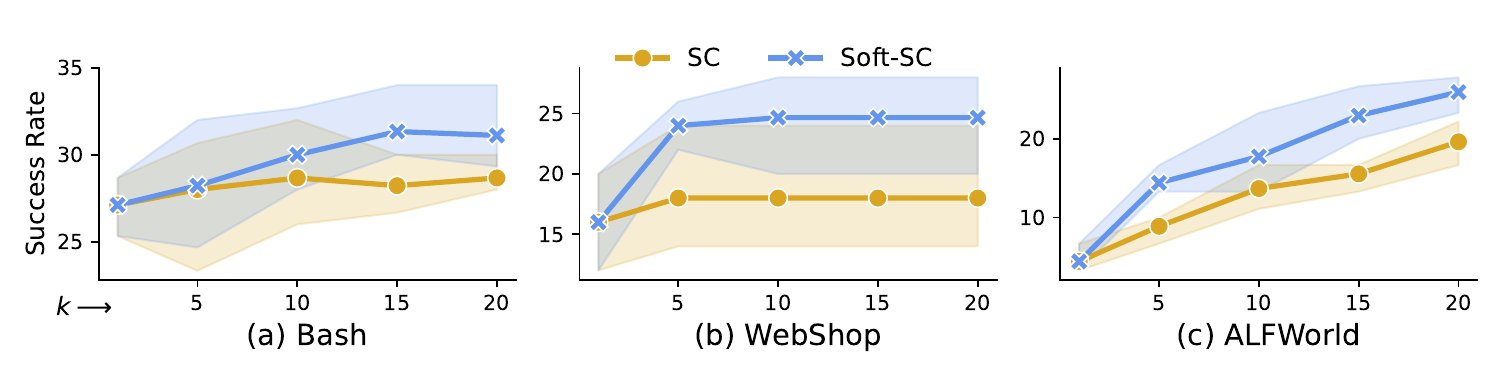}
    \caption{ 
    Compared to self-consistency (SC), our method \method{}, exhibits better scaling with respect to the number of samples $k$, generally outperforming SC for each $k$. We use CodeLlama-34B~\cite{roziere.b.2023codellama} to compute success rates on the test set of Bash and WebShop. Due to computational cost, for ALFWorld we use Mistral-7B~\cite{jiang2023mistral} on a 30-task subset of the test set. 
    }
    \vspace{-1em}
    \label{fig:intro_web}
\end{figure*}

The performance of large language models (LLMs) can be greatly improved by generating multiple samples and scoring their answers before making a final selection. 
One popular and effective ``sample and select'' approach is \emph{Self-Consistency} \citep[SC;][]{wang2023self}, which leverages chain-of-thought prompting~\cite{wei2022chain} to generate multiple solutions for each input query and then determines the final answer via a majority vote. 
While SC has demonstrated consistent benefits on question-answering datasets, we find it provides minimal gains in several interactive settings where LLMs act as agents to generate a sequence of actions.
SC's selection mechanism relies on \emph{exact match} in order to tally votes, i.e., it scores answers based on their frequency.
However, in interactive domains, multiple distinct and valid answers -- in this case, actions -- can be generated at each step. This diminishes the effectiveness of SC over actions because the likelihood of generating identical actions decreases as the number of plausible options grows.
For instance, a model tasked with predicting bash commands based on user queries has a very large action space (all bash commands) and could generate semantically equivalent commands that differ in their surface form (e.g., \texttt{ls -ltr} vs \texttt{ls -trl}).\footnote{For Bash program prediction with five samples, SC fails to produce a single majority action $86\%$ of the time.}
Therefore, deriving a signal from voting in LLM-agent domains would require sampling a large number of actions at each step throughout a lengthy trajectory, reducing efficiency and making SC prohibitively expensive (cf. \cref{fig:intro_web}).

We hypothesize that relaxing the strict scoring criterion from votes tallied by exact match to a continuous score will address the shortcomings of SC in two ways: (i) improving \emph{task performance} in sparse action spaces; and (ii) increasing \emph{sample efficiency}, i.e., higher success rates with fewer samples.
We propose \emph{\fullform{}} (\method), a continuous relaxation of exact-match sample and select methods.
Unlike match-based voting, \method{} handles cases without a \emph{unique} majority answer. 
Crucially, for a white-box model, \method{} incurs no additional cost and requires no external tests or metrics, as the probabilities used are already produced. 
Finally, we show that \method{} can be used to rescore black-box models' outputs and can be integrated into an efficient variant of SC.

We test \method{} on three diverse interactive domains: Bash~\cite{NEURIPS2023_Intercode}, WebShop~\cite{NEURIPS2022_webshop}, and ALFWorld~\cite{sridhar2021alfworld}.

\noindent\textbf{Summary of Key Findings:}  
\begin{enumerate}[leftmargin=*,noitemsep,nolistsep]
\item We demonstrate that \emph{\method{} outperforms SC} with the same number of samples, e.g., by up to $6.6\%$ on WebShop using CodeLlama-34B.  

\item \method{} exhibits better sample efficiency i.e.,  produces better performance than SC with fewer samples (cf. \cref{fig:intro_web}). 

\item \method{} scales better with model size than SC, increasing performance by $8.8\%$ on Bash as model size increases from 7B to 70B, as opposed to only $5.8\%$ improvement by SC. 

\item \method{} can be combined with smaller LMs to \textit{score generations from black-box models}. We observe that \method{} outperforms SC on closed-source models such as GPT-4~\cite{achiam2023gpt} by up to $4\%$ on WebShop. 
\end{enumerate}

\section{Methodology}
\subsection{\fullform{} (\method{})}
\label{sec:method}
Following \citet{wang2023self}, for a given input $\mathbf{x}$ containing the task description, we generate $k$ solutions
using temperature-based sampling~\cite{ackley1985learning, ficler2017controlling}. 
To perform selection, we score the action $\mathbf{y}_i$ resulting from each solution using the aggregated probability of the action's tokens.  
For an action $\mathbf{y}$ composed of tokens $y_1, \ldots, y_n$, we define $\mathrm{score}(\mathbf{y}) = f\big(\{P_{\textrm{LM}}(y_i | y_{<i}, \mathbf{x}) \:\forall i \in [1, n] \}\big)$ where $f\in \{\textrm{min}, \textrm{mean}, \textrm{product}\}$. We choose the aggregation method based on dev set performance.
We use mean probability for Bash and ALFWorld and min probability for Webshop.  
We then choose an action $\hat{\mathbf{y}}$ with the highest score, i.e., $\mathbf{\hat{y}} = \argmax_{j=1}^{k} \:\mathrm{score}(\mathbf{y}_j)$. 
Further details and results for $f$ options are provided in \cref{append:other_aggs}.

\subsection{Adaptive \fullform{}}\label{sec:adaptive}
To improve efficiency, \citet{aggarwal-etal-2023-lets} introduce adaptive-consistency, which reduces the number of samples ($k$) by approximating the final vote tally per example via sampling discrete vote distributions from a prior and stopping when the samples converge.
Instead of sampling from discrete distributions, we choose $k$ by aggregating likelihood scores until a score threshold $\tau$ is reached. 
Following \citet{stengel-eskin-van-durme-2023-mean}, we use the minimum probability for comparing with the threshold. 
We sample one action at a time, stopping when  $\sum_{j=1}^k \text{min}_{i=1}^{|\mathbf{y}_{\boldsymbol{j}}|} P_{\textrm{LM}}({y}_i | {y}_{<i},\mathbf{x}) \geq \tau$, where
$\tau$ is chosen on the dev set (cf. \cref{app:adaptive}).

\begin{table*}[t]
\centering
\small
\begin{tabular}{lccccc}
\toprule
\multirow{2}{*}{\textbf{Method}} & \multirow{2}{*}{\bf \# Samples ($\boldsymbol k$)} & \textbf{Bash} & \multicolumn{2}{c}{\textbf{WebShop}} & \textbf{ALFWorld} \\
\cmidrule{3-6}
&  & \bf SR & \bf Score & \bf SR & \bf SR \\
\midrule
Greedy decoding & 1 & $27.1\interval{1.7}$ &$33.1\interval{2.8}$ & $16.0\interval{4.0}$ &  $18.7\interval{2.1}$ \\
Self-Consistency \citep{wang2023self} & 10 & $28.7\interval{3.1}$ & $36.4\interval{3.3}$ & $18.0\interval{5.3}$ & $20.5\interval{2.9}$ \\
Adaptive-Consistency \citep{aggarwal-etal-2023-lets} & [5.0, 7.3]$^\dagger$ & $27.3\interval{2.4}$ & $38.8\interval{2.4}$ & $19.3\interval{4.2}$ & $20.8\interval{3.2}$ \\ 
\midrule
\method{}  & 5 & $28.2\interval{3.7}$ & $44.2\interval{3.8}$ & $24.0\interval{2.0}$ & $22.7\interval{2.5}$ \\
\method{} & 10 & $\mathbf{30.0}\interval{2.4}$ & $\mathbf{46.0}\interval{6.0}$ & $\mathbf{24.6}\interval{4.2}$ & $\mathbf{25.2}\interval{3.2}$ \\
 Adaptive \method{} & [5.0, 5.9]$^\dagger$ & $\mathbf{30.0}\interval{2.7}$  & $44.5\interval{4.1}$ & $23.3\interval{2.3}$ & $23.9\interval{2.9}$ \\ 
\bottomrule
\end{tabular}
\caption{Success rates and scores from CodeLlama-34B, averaged across three seeds ($\pm$ standard deviation). With a fixed $k=10$, \method{} outperforms self-consistency by an average of $4.2\%$, across datasets.
Adaptive sampling uses fewer samples on average than adaptive-consistency while also increasing performance. \\
$^\dagger$Adaptive methods result in differing average $k$ for each dataset, range reported here.  }
\label{tab:main}
\end{table*}

\subsection{Datasets}
We test on three representative English LLM agent datasets; further details can be found in \cref{app:bash,app:web,app:alf}.

\paragraph{Bash.} We use \citet{NEURIPS2023_Intercode}'s bash data, which consists of 200 user queries or instructions that can be completed via $\tt{bash}$ actions. We split these into 50 dev and 150 test. 
The agent's performance is measured via success rate. 
Bash represents a domain with a large action space, as the space of possible bash commands is very large, and many of the queries involve stringing multiple functionalities together into a complex command.

\paragraph{WebShop.} WebShop~\citep{NEURIPS2022_webshop} is a simulated online shopping website environment. 
Success is measured both by a score $\in [0,1]$ reflecting how well the purchased product matches the user's criteria; the success rate is the rate of perfect scores. Following \citet{zhou2023language}, we report performance on a subset of 50 user queries.
WebShop also has a large action space, as there are $1.18$ million real-world products to select from.

\paragraph{ALFWorld.} ALFWorld~\citep{sridhar2021alfworld} is a text-game adaption~\cite{cote2019textworld} of the embodied ALFRED benchmark~\cite{shridhar2020alfred} in which an agent performs household chores (e.g., cleaning a mug) via a series of low-level actions. 
We evaluate on 134 unseen tasks and report the overall success rate.
ALFWorld requires agents to generate long action sequences, involving thousands of valid actions at each step for some tasks.

\paragraph{Metrics.}
All these interactive tasks provide a goal and associated environments to execute the LLM-generated actions to accomplish said goal. After executing each action, the environment returns the observation and reward. The observation is a natural language description of the state of the system after executing the action, and the reward indicates if the goal was successfully achieved. The reward can be used to obtain a \emph{success rate}, the percentage of examples with the maximum reward possible. Further details on the rewards for each domain can be found in \cref{app:bash,app:web,app:alf}.

\subsection{Baselines}
We compare \method{} against the following: 

\paragraph{Greedy Decoding.} We sample a single solution with greedy decoding on all datasets; all prompts are given in \cref{append:prompts}.
This is equivalent to both SC and \method{} when $k\!=\!1$, as no selection is needed for a single sample. 

\paragraph{Self-Consistency (SC).} We use self-consistency as described by \citet{wang2023self}, with majority voting as the selection criterion. We tally votes towards each response using exact match. 

\paragraph{Adaptive-Consistency (AC).} As described in \cref{sec:adaptive}, 
\citet{aggarwal-etal-2023-lets} introduce an adaptive version of SC that improves efficiency by 
adaptively reducing the number of samples.
We implement their Beta estimator for all of our settings.  
Further details can be found in \cref{app:adaptive}.

\section{Results and Discussion}
Unless mentioned otherwise, we report average performance on 3 random seeds for each test set.

\paragraph{For the same number of samples $k$, \method{} outperforms SC.}
\label{ssec:rq1}
In \cref{tab:main}, we compare \method{} against the baselines on all datasets using CodeLlama-34B on the test sets. 
While both SC and \method{} boost performance over the greedy decoding baseline, we find \method{} results in a larger margin of improvement, $8.6\%$ on WebShop  (SC only yields $2\%$). 
For the same number of samples ($k\!=\!10$), \method{} outperforms SC by $1.3\%$, $6.6\%$, and $4.7\%$ (success rate) on Bash, WebShop, and ALFWorld respectively.  
Comparing the adaptive version of \method{} with \citet{aggarwal-etal-2023-lets}, our likelihood-based scores not only improve efficiency by generally using fewer samples, but \emph{also} outperforms AC, e.g., by $4\%$ on WebShop and $3.1\%$ on ALFWorld. 

\paragraph{\method{} exhibits better scaling with $\boldsymbol{k}$.}
\label{ssec:rq2}
In \cref{tab:main}, even with $k\!=\!5$, \method{} can outperform SC with $k\!=\!10$, e.g., with $2.2\%$ improvement on ALFWorld.
In \cref{fig:intro_web}, we compare this trend across more values of $k$, showing the scaling of \method{} and SC with an increasing $k$. 
We observe that SC provides minimal gains even as $k$ increases, e.g., on Bash increasing $k$ from 5 to 20 only yields $1\%$ point improvement in success rate.
On the other hand, \method{} consistently improves success rates with $\sim\!3\%$ points improvement as $k$ goes from $5$ to $20$. While SC does improve the success rate of Mistral-7B on ALFWorld with increasing $k$, \method{} yields greater performance gains using fewer samples, e.g., \method{} with $k\!=\!5$ is comparable to SC with $k\!=\!10$.

\begin{figure}[t]
    \centering
    \includegraphics[trim={0.2cm, 0.35cm, 0.1cm, 0.8cm},clip,width=\linewidth]{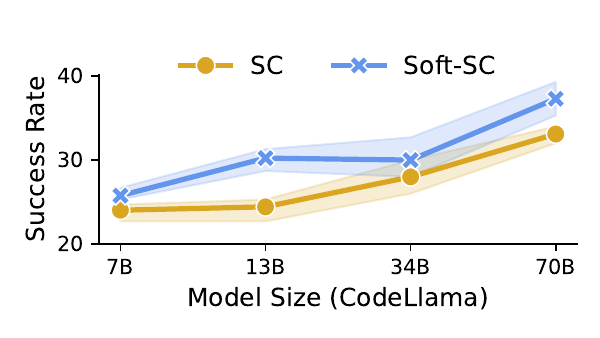}
    \vspace{-1.5em}
    \caption{Scaling with model size on Bash (test). 
    \method{} improves over SC for all model sizes.}
    \label{fig:scaling}
    \vspace{-0.5em}
\end{figure}

\begin{figure}[t]
\centering
\includegraphics[trim={0.15cm, 0.22cm, 0.15cm, 0.15cm},clip,width=\linewidth]{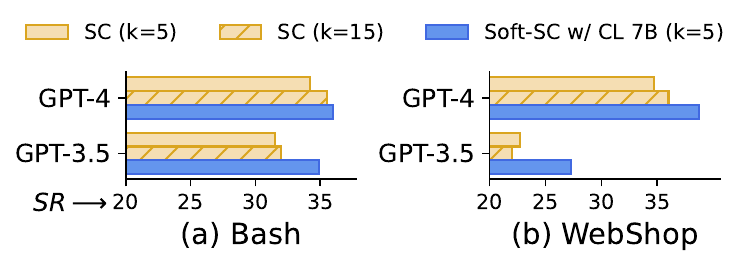}
\vspace{-1em}
    \caption{\method{} can be used to score outputs from black-box models on Bash and Webshop (test), improving success rate (SR) over self-consistency. }
    % \vspace{-0.5em}
    \label{tab:blackbox}
\end{figure}

\paragraph{\method{} effectively scales with model size.}
As we scale up the size of the LM, we find that \method{} continues to provide improvements over SC. 
\cref{fig:scaling} shows the scaling trends for CodeLlama models ranging from 7B to 70B parameters on Bash and WebShop with a fixed $k\!=\!10$.  
For each LM, \method{} always outperforms SC. 
Furthermore, \method{} often allows smaller LMs to outperform larger members of the same model class, e.g., CodeLlama-13B with \method{} outperforms CodeLlama-34B with SC. 
This points to additional efficiency gains from \method{}, as it can allow smaller models to replace larger ones. 
\paragraph{\method{} improves black-box models more than SC.}\label{ssec:rq4}
\method{} requires access to token probabilities to score actions. 
However, the most performant LLMs are typically black-box models, often with limited or no access to logits~\cite{achiam2023gpt, pichaiimportant, anthropic}.
In \cref{tab:blackbox}, we study whether (smaller) open-source LMs can be used to score generations from GPT-3.5 and GPT-4.
Here, we observe that \method{} offers improvements over SC for a given black-box model, e.g., $4\%$ for GPT-4 on WebShop and $1.8\%$ on Bash when \method{} uses the \emph{same} number of generations from the black-box models as SC.  Furthermore, even though Soft-SC requires 2 model calls (one to the black-box model and one to a smaller open-source model), \method{} with $k = 5$ (total 10 calls) outperforms SC with $k = 15$ (total 15 calls to the black-box LLM), which shows that our method is significantly more efficient and effective since it can achieve better performance with fewer calls. Note that half of the calls for \method{} are to a 7B model, likely making them much less expensive than calls to the black-box model.

\vspace{-0.5em}
\paragraph{Calibration is not required for strong \method{} performance.}
Given that \method{} selects options using scores based on token probabilities, we investigate whether a model has to be well-calibrated for \method{} to work.  
We compute the correlations between two standard calibration metrics -- ECE \citep{naeini.m.2015} and AUROC  -- and absolute \method{} performance for CodeLlama-34B across seeds and values of $k$ on WebShop and Bash test sets.
The full plot is shown in \cref{append:calibration}. 
We find a moderate negative correlation with AUROC ($r\!=\!-0.55$) on Bash and no significant correlation on WebShop); there is no significant correlation for ECE. 
In other words, having a well-calibrated model is \emph{not} a prerequisite for \method{}.
This may be because calibration metrics do not measure \emph{ranking} performance, which is central to our approach.

\begin{table}[t]
    \centering
    \small
    \begin{tabular}{lccc}
    \toprule
    \textbf{k} & \textbf{SC} & \textbf{\method{} (logit)} & \textbf{\method{} (verb.)} \\ 
    \midrule
    5 & $28.0\interval{4.1}$ & $\mathbf{28.2}\interval{3.7}$ & $27.8\interval{2.2}$ \\
    10 & $28.7\interval{3.1}$ & $\mathbf{30.0}\interval{2.4}$ & $27.6\interval{2.0}$ \\
    \bottomrule
    \end{tabular}
    \vspace{-0.5em}
    \caption{Success rates for CodeLLama-34B on Bash with logit-based confidence vs. verbalized (verb.) confidence, averaged across three seeds ($\pm$ std. dev.).}
    \label{tab:verbalized}
    \vspace{-0.5em}
\end{table}
\vspace{-0.5em}
\paragraph{Logit-based score outperforms verbalized confidence score.}
Recent work has explored prompting language models to express uncertainty or confidence score in human language \citep{lin2022teaching, tian-etal-2023-just, xiong2024can}. We study whether verbalized confidence scores can be used for selection instead of logit-based scores. We follow  \citet{lin2022teaching} in prompting models to generate verbalized scores, which we then use for selection. As shown in \cref{tab:verbalized}, verbalized scores perform poorly when used in place of logit-based scores on Bash:
Soft-SC with logits outperforms the verbalized method by 2.4\% with $k = 10$.

\section{Related Work}
\paragraph{Sample and Select Methods for LLMs.}
Ensembling via voting over or aggregating outputs \citep{breiman1996bagging, freund1997decision} can improve a classifier's performance. \citet{wang2023self} apply this paradigm to improve LLMs on reasoning tasks, introducing self-consistency (SC).
We find that the majority voting used in SC is not suited for LLM-agent domains because the LLM's generations may not exactly match when the action space is large.
\citet{chen2023universal} generalize SC by prompting the LLM to determine consistency. However, LLMs still struggle to determine consistency in interactive domains where the task is partially observable~\cite{ruan2023toolemu}.
In contrast to \method{}, past work examining re-ranking strategies in code generation~\cite{chen2022codet, li2024using} or reasoning \citep{golovneva2022roscoe, prasad-etal-2023-receval} rely on external test cases or model-based metrics to score responses.

\paragraph{LLM-Agents.}
LLMs have proven to be effective agents across a diverse array of multi-step tasks, e.g., mathematical reasoning~\cite{wei2022chain}, tool-usage~\cite{schick2023toolformer, qin2023toolllm}, robotic navigation~\cite{ahn2022can, singh2023progprompt}, and code-generation~\cite{NEURIPS2023_Intercode}. 
Standard LLM-agent solutions employ chain of thought prompting~\cite{wei2022chain} interleaved with permissible actions within an environment~\cite{yao2023react}. 
Several follow-up works improve upon this pipeline by building feedback over multiple trials~\cite{shinn2023reflexion}, decomposing tasks~\cite{prasad2023adapt}, or searching over trajectories~\cite{yao2023tree}. 
\method{} is complementary to these approaches, which can be seen as improvements to CoT for a single generation.
Note that our work focuses on a single LLM agent~\cite{andreas-2022-language} interacting with an external environment to accomplish tasks; this single agent is compatible with other lines of work on discussion among multiple LLM agents~\cite{du2023improving,chen2023reconcile}.

\section{Conclusion}
After establishing the shortcomings of standard voting-based SC in interactive tasks, we introduced \method{}, which relaxes the exact-match scoring function used by SC to a continuous score.
On three commonly used interactive benchmarks, we showed that \method{} results in improved performance and increased efficiency.
We also show that \method{} is compatible with both white-box and black-box models and that it can be integrated into a more efficient adaptive variant of self-consistency. 
Finally, we find that a well-calibrated model is not required for \method{} to work well, and that logits outperform verbalized confidence scores.

\section{Limitations and Broader Impacts}
\paragraph{Limitations.} In \cref{sec:intro}, we pointed out that excessive diversity can lead to failures for SC, as no majority will emerge. 
However, both SC and \method{} rely on some amount of output diversity: if the model generates $k$ identical samples, then the output will be no better than generating one. 
One major motivation for \method{} is efficiency; \method{} substantially improves performance and is able to do so with fewer samples than SC, but it still requires multiple samples from an LLM.
Thus, like all sample and select methods, \method{} has a greater cost than greedy decoding. 
In \cref{ssec:rq4}, we demonstrate that \method{} can be used to rerank outputs from other models that do not consistently provide logits.
While \method{} shows major improvements in reranking the outputs of black-box models, it could be applied directly without a smaller scoring model if the generation model's underlying logits (which exist by design) were made accessible to users. 

\paragraph{Broader Impacts.} Large language models have the potential for negative applications and malicious use~\cite{weidinger2021ethical,bommasani2021opportunities}. Our work improves LLM performance, meaning it could also be negatively applied. 
As our work is applied to LLMs operating as agents, it shares the inherent risk of all LLM agent work, namely that the LLM agent could potentially make mistakes and that its actions could lead to negative outcomes for the user. 
Overall, we believe this risk is mitigated by our use of simulated benchmarks (i.e., no agent we evaluate or develop can affect the world) and by the fact that our work improves agent accuracy, making adverse outcomes less likely. 

\section*{Acknowledgements}
We thank Justin Chen and Swarnadeep Saha for their valuable help and feedback on the paper. This work was supported by NSF-AI Engage Institute DRL-2112635, DARPA Machine Commonsense (MCS) Grant N66001-19-2-4031, and the Accelerate Foundation Models Research
program. The views contained in this article are those of the authors and not of the funding agencies. 

\bibliography{anthology,custom}

\appendix

\section{Method and Dataset Details}
\label{append:method}
\subsection{Hyperparameters} 
\label{append:hyperparams}
We select the threshold $\tau$ on the dev set for both Adaptive-Consistency baseline and Adaptive \method{}. 
For Adaptive-Consistency baseline, we set the threshold $\tau$ of 0.8, 0.85, and 0.8 for Bash, WebShop, and ALFWorld respectively. 
For Adaptive \method{}, we set the threshold $\tau$ to 0.95, 3.0, and 3.5 for Bash, WebShop, and ALFWorld respectively. 
Because Adaptive \method{} accumulates minimum probabilities over $k$ samples for comparing with the threshold, the threshold may be $\geq 1$. 

For greedy decoding, we use a temperature of 0.7 for all datasets. In case of sampling $k > 1$ outputs from the model, we set the temperature of open-source models to 0.7 for Bash, 0.9 for WebShop, and 0.9 for ALFWorld, with \texttt{top-p} value of 0.9 and \texttt{top-k} value of 40, and with \texttt{max\_tokens} set to $100$. For obtaining generations from the OpenAI API, we use a temperature of 0.7 for Bash, 0.9 for WebShop and ALFWorld and \texttt{top-p} value of 1 for all datasets.

\subsection{Model Checkpoints and Licenses}
Webshop, Bash, and ALFWorld all have MIT licenses.
CodeLlama is released under a custom permissive license available here: \url{https://github.com/facebookresearch/llama/blob/main/LICENSE}.
Mistral uses an Apache License 2.0.
For CodeLlama, we used the \texttt{CodeLlama-*b-Instruct} checkpoints.  For Mistral, we used the \texttt{Mistral-7B-Instruct-v0.2} checkpoint.
All open-source models were accessed via Huggingface Transformers \citep{wolf2019huggingface}.
For OpenAI models, we used the \texttt{gpt-3.5-turbo-0613} and \texttt{gpt-4} checkpoints.
All models were run for inference only with \texttt{int-8} quantization on Nvidia 40GB A100 GPUs. 
We will release our code under an MIT license.

\subsection{Bash}
\label{app:bash}
\citet{NEURIPS2023_Intercode} propose an interactive benchmark for evaluating LMs on a bash coding task, created by bootstrapping queries from NLP2Bash benchmark~\cite{lin-etal-2018-nl2bash}. The dataset has 200 user queries or instructions that can be completed via $\tt{bash}$ actions, which we split into 50 dev and 150 test. 
After each action is executed, the agent observes the corresponding output from the file system. The agent's performance is measured via success rate, which is determined by a reward function based on modifications to the file system with respect to a gold command as well the latest execution output -- a success means the reward is $1.0$. For example, given a query \textit{"find files in the /workspace directory and sub-directories, that changed within last hour"}, the agent generates a corresponding command \texttt{find /workspace -cmin -60}. 

\paragraph{Setup.}
We focus on the single-turn setting instead of the multi-turn setting because we find the observation (i.e., the execution output of the action) from the Bash environment and the oracle reward rarely helps the agent generate correct commands.  
In our preliminary experiments, we observed that generating multiple commands using temperature-based sampling under the single-turn setting resulted in a success rate comparable to or even better than the multi-turn setting. 
Furthermore, in real-world scenarios, it is impossible to obtain oracle rewards to determine whether the generated commands are correct.
Therefore, we prompt the LLM with a simple description of the task setting to sample $k$ commands that would address the query. 
The final command selected by different methods is executed in the InterCode Bash environment and the response is scored to get the success rate.

\paragraph{Metric.}
After submitting the generated action, the environment returns a reward $r \in [0, 1]$.
The reward function takes into account the differences in the file system resulting from executing the predicted command and the file system resulting from executing the gold command, as well as the latest execution output. The \emph{Success Rate} (SR) metric is defined as the proportion of tasks where $r = 1$.

\subsection{WebShop}
\label{app:web}
WebShop~\citep{NEURIPS2022_webshop} is a simulated online shopping website environment with 1.18 million real-world products.  The underlying task requires an agent to navigate a simulation of a shopping website via a series of commands and buy a suitable product as per the user's instruction (e.g., 3oz bottle of natural citrus deodorant for sensitive skin under \$30). At the end of the trajectory, the environment returns a numeric score $\in [0,1]$ reflecting the degree to which the bought product matches the input criteria. Performance is measured based on the score as well as the success rate (i.e., a perfect score of 1).
WebShop also has a large action space, as there are millions of products to select from.
 We use 30 user queries \emph{not} in the test set to finalize our prompts and thresholds used for adaptive consistency as well as adaptive \method{}.

\paragraph{Setup.} Following \citet{prasad2023adapt}, we factorize the underlying agent into two modules: (i) selecting a suitable product, and (ii) buying a selected product. This simulates a ``cart'' functionality in online shopping. Given a user query, the agent first employs the search functionality and picks a few relevant products from the search page. It then explores the corresponding product page, matches its features, and determines if it can be added to the cart. We prompt the LLM to generate $k$ such trajectories, potentially adding up to $k$ products to the cart. In the end, we select a product by majority vote over product IDs and use a separate prompt to get the agent to buy the product while selecting relevant product options such as color, size, etc. The corresponding prompts are shown in \cref{append:prompts}.

Note that due to the discrete and discontinuous nature of exact match~\cite{schaeffer2023emergent}, SC can only perform selection over products.
Given a description, SC navigates through the environment and selects multiple product pages, indexed by their IDs; these IDs can be aggregated via voting.
However, within each product page, there are numerous follow-up options that must be selected, and which cannot be voted on as their selection happens across multi-step trajectories. 
Once a majority product is selected, SC uses a greedy action trajectory based on ReAct~\citep{yao2023react} to specify the options for a selected product; this often results in suboptimal products being bought, as SC often picks the default option. 

In contrast, the scoring criterion in \method{} allows us to score and select from trajectories to first select products as well as to specify their options and buy them, generating and scoring $k$ trajectories overall. 
Thus, \method{} accounts for diversity in each stage and yields higher performance. 
For example, for the user query \emph{``natural looking long clip in extensions under \$40''} SC tallies votes for products IDs the cart after the product selecting stage: [\texttt{B09QQLDJ93}, \texttt{B093BKWHFK}, \texttt{B09QQLDJ93}], picking the \texttt{B09QQLDJ93} as it forms a majority. 
It then uses a greedy ReAct trajectory to select the final options (e.g., the color) and to buy the item.  
\method{}, on the other hand, can differentiate between action trajectories sampled for buying the \emph{same} product ID, allowing it to distinguish between a final selection that has the default color ``pink'' and the correct product that uses the color ``brown'' -- resulting in different scores from the environment.

\paragraph{Metric.}
When the LLM agent generates a \textbf{buy} action at the end of the trajectory, the environment returns a reward $r \in [0, 1]$ reflecting the degree to which the bought product matches the input criteria. The \emph{Success Rate} metric is defined as the portion of tasks where $r = 1$. The \emph{Score} metric is defined as ($100 \times$ avg. reward), which captures the average reward obtained across different task trajectories.

\subsection{ALFWorld}
\label{app:alf}

ALFWorld~\citep{sridhar2021alfworld} is a text-game adaption~\cite{cote2019textworld} of the embodied ALFRED benchmark~\cite{shridhar2020alfred}. The underlying task requires the agent to perform basic household chores such as finding a mug, cleaning it, and putting it on a countertop via a series of low-level actions (e.g., ``go to sink''). After each action, the environment provides textual feedback (e.g., the contents of the cabinet after it is opened). We evaluate on 134 unseen tasks spanning 6 task types and report the overall success rate. In \cref{fig:intro_web}, due to computational requirements of using a larger number of samples, we report performance on a subset of the test split consisting of a total of 30 tasks, picking 5 from each task type. For the dev set, we use a disjoint set of 12 tasks from the `valid seen' split of ALFWorld. This is only used to select the scoring criteria, e.g., mean, min, or product, and the thresholds for the adaptive variants.

\paragraph{Setup.} Unlike WebShop, tasks in ALFWorld cannot be decomposed uniformly such that each sub-task is handled by an independent agent without significant planning and communication overhead~\cite{prasad2023adapt}. For instance, the sub-tasks involved in ``putting a clean mug on a countertop'' vary considerably from the sub-tasks involved in ``examining a spray-bottle under a desklamp''. Therefore, in ALFWorld, at each step, we sample $k$ actions, and for SC perform majority voting over these $k$ actions.  Note that both \method{} and SC only score \emph{actions}, not thoughts or comments generated by the agent to aid in problem-solving. We continue sampling responses until a valid action is reached, skipping ``thought'' actions (i.e., generations starting with ``Think:'') as well as comments. We only allow the selection of actions, ignoring the reasoning generated before the action. 
Note that both SC and \method{} are more computationally demanding in the case of ALFWorld, since we perform selection over actions at each step, as compared to WebShop, where selection is performed once at the end of the selection phase over products. 
Following \citet{yao2023react}, the prompt to the LLM includes one in-context trajectory corresponding to a query from the same task type as the test instance.

\paragraph{Metric.}
After each action generated by the LLM agent, the environment provides textual feedback (e.g., the contents of the cabinet after it is opened). The feedback \emph{``You won!''} in addition to reward $r = 1$ indicates that the agent has completed the task successfully. The \emph{Success Rate} metric is the percentage of tasks where the agent succeeds.

\subsection{Aggregation Methods}
\label{append:other_aggs}
For a given input $\mathbf{x}$ containing the task description and a corresponding sampled action $\mathbf{y}$ composed of tokens $y_1, \cdots, y_n$, we can compute $\mathrm{score}(\mathbf{y})$ using the following probability aggregation methods:
\begin{itemize}[leftmargin=*,noitemsep,nolistsep]
    \item \textbf{Mean}: $\mathrm{score}(\mathbf{y}) = \frac{1}{n}\sum\limits_{i=1}^nP_{\textrm{LM}}(y_i | y_{<i}, \mathbf{x})$
    \item \textbf{Min}: $\mathrm{score}(\mathbf{y}) = \mathop{\min}\limits_{1\leq i\leq n}P_{\textrm{LM}}(y_i | y_{<i}, \mathbf{x})$
    \item \textbf{Length-Normalized Product}: $\mathrm{score}(\mathbf{y}) = \exp\left(\frac{1}{n} \sum_{i=1}^n \log P_{\textrm{LM}}(y_i | y_{<i}, \mathbf{x})\right)$.
\end{itemize}
\vspace{0.5em}
For Bash and ALFWorld, we perform scoring and selection at the action level, where the mean probability serves as an effective measure of the overall confidence in an action being the correct response to a given query. WebShop involves trajectory-level evaluations, where the correctness of a sequence of actions (a trajectory) towards accomplishing a task is assessed. In the case of WebShop, the trajectory represents a sequence of actions to \emph{select} a suitable product based on the user query by navigating through a series of webpages; this sequential nature makes min better-suited. We also demonstrate experimental results on dev set for all aggregation methods to validate our explanation in \cref{tab:my_label}.

\begin{figure*}[t]
	\centering
	\begin{minipage}[b]{0.49\textwidth}
		\subfigure[Bash]{
			\includegraphics[width=0.5\textwidth]{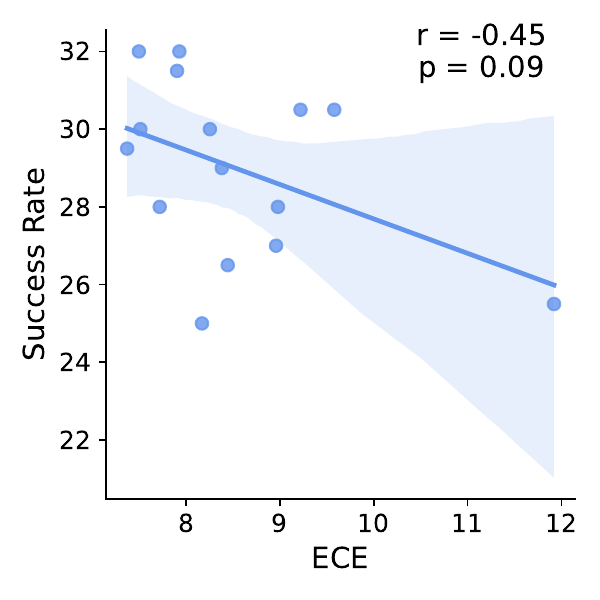} 
			\includegraphics[width=0.5\textwidth]{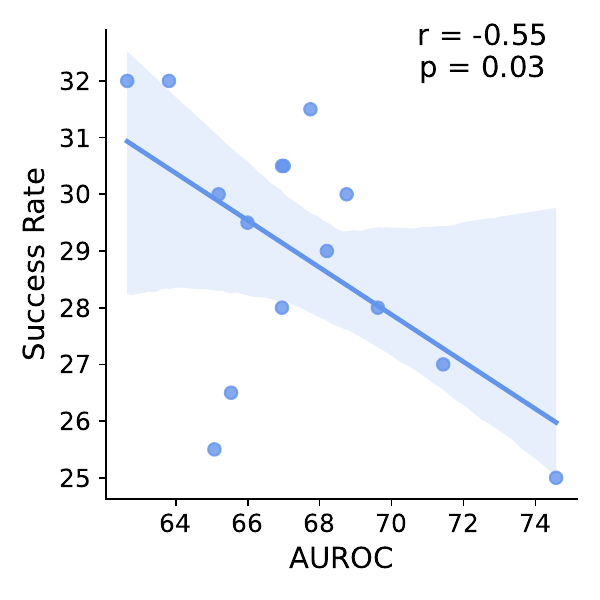}
			\label{fig:bash_calibration}
	}
	\end{minipage}
	\begin{minipage}[b]{0.49\textwidth}
		\subfigure[Webshop]{
			\includegraphics[width=0.5\textwidth]{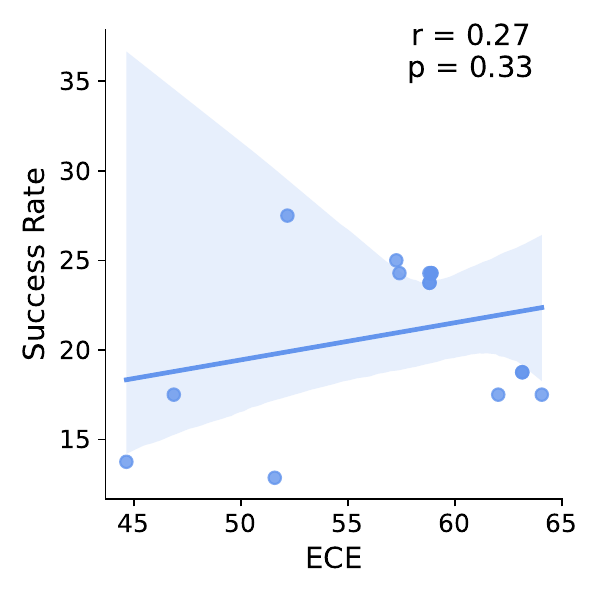} 
			\includegraphics[width=0.5\textwidth]{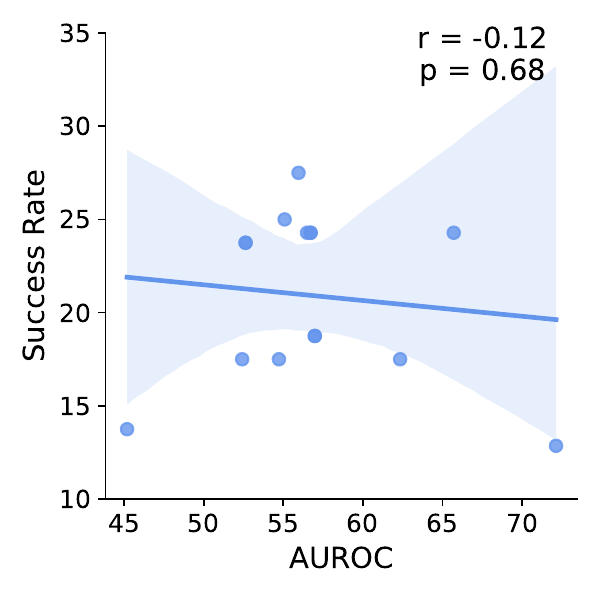}
			\label{fig:webshop_calibration}
	}
	\end{minipage}
 \vspace{-1em}
	\caption{The Pearson correlations between two standard calibration metrics -- ECE and AUROC -- and \method{} performance for CodeLlama-34B across seeds and values of $k$ on Bash and Webshop test set.}
	\label{fig:calibration}
 \vspace{-0.5em}
\end{figure*}

\begin{table}[t]
    \centering
    \small
    \begin{tabular}{lccc}
    \toprule
    \textbf{Method} & \textbf{Bash} & \textbf{WebShop} & \textbf{ALFWorld} \\ 
    \midrule
    SC & 20.0 & 22.0 & 6.70 \\
    min & 18.0 & \textbf{33.0} & 10.0 \\
    mean & \textbf{24.0} & 30.0 & \textbf{16.7}\\
    product & 22.0 & 16.7 & 13.3\\ 
    \bottomrule
    \end{tabular}
    \caption{Dev success rates for one seed across aggregation methods. 
    For Bash and WebShop we use CodeLlama-34B and for ALFWorld we use Mistral-7B.}
    \label{tab:my_label}
\end{table}

\subsection{Baselines}
\paragraph{Greedy Decoding.}
We sample trajectories
with greedy decoding on all datasets; prompts are given in \cref{append:prompts}.
For WebShop and ALFWorld, we follow a ReAct prompt format \citep{yao2023react} while for Bash we follow the standard format provided by \citet{NEURIPS2023_Intercode}.
This is equivalent to both SC or \method{} when $k=1$ (since with a single sample, there is no selection needed, making the selection strategy irrelevant). 
\paragraph{Self-Consistency (SC).} We use self-consistency as described by \citet{wang2023self}, with majority voting as the selection criterion. We tally multiple votes towards a response only if the model generates the \emph{exact} response multiple times.

\subsection{Adaptive \method{}}
\label{app:adaptive}
To improve sample efficiency, \citet{aggarwal-etal-2023-lets} introduce adaptive-consistency (AC), which reduces the number of samples ($k$) needed for selection by approximating the final vote tally through sampling.
Specifically, AC adds generations one at a time (i.e., it increments $k$ starting from 1) and terminates when a stopping criterion is satisfied or the number of generations has reached the maximum allowed. 
The stopping criterion is based on samples from a discrete distribution over vote distributions, parameterized by the current vote counts; these samples represent likely future vote distributions given the current trends. 
If the samples have converged, then further generations are unnecessary.
For example, if $5/10$ samples have been generated and $4$ are identical, then the probability that the next $5$ will change the majority vote is vanishingly small, meaning that generating further solutions is wasteful. 
On the other hand, if there is no clear majority winner after $5$ samples, further solutions would be needed.

We can apply a similar methodology to \method{}.
However, instead of estimating $k$ by sampling from a discrete vote distribution, we estimate the stopping criterion for sampling by aggregating likelihood scores until a sufficient score threshold $\tau$ is reached. 
While we use average probability across tokens for selection, we find that this score is poorly calibrated.
Following \citet{stengel-eskin-van-durme-2023-calibrated}, who found minimum token probabilities to be better calibrated, we use the minimum probability for comparing with the threshold. Therefore, we sample actions one-at-a-time and stop when the number of samples $k$ is such that  $\sum_{j=1}^k \text{min}_{i=1}^{|\mathbf{y}_{\boldsymbol{j}}|} P_\theta({y}_i | {y}_{<i},\mathbf{x}) \geq \tau$.
The threshold $\tau$ is a domain-specific hyperparameter that we select based on a dev set (discussed in \cref{append:hyperparams}). 
Specifically, we set the threshold $\tau$ to 0.95, 3.0, and 3.5 for Bash, WebShop, and ALFWorld respectively.
Note that in this case, the threshold can be $>1$ as it represents a threshold on cumulative confidence values, rather a threshold on true probability distribution.
This differs from adaptive-consistency, for which the threshold is over a normalized probability, i.e., it must be less than $\leq 1$.

\vspace{-0.5em}
\section{Calibration}
\vspace{-0.5em}
Following past work \citep{kuhn2023semantic, stengel-eskin-van-durme-2023-calibrated}, we use Expected Calibration Error (ECE) and Area Under the Receiver Operator Characteristic curve (AUROC) to check the calibration of scores used in \method{}:

\paragraph{Expected Calibration Error (ECE)} \citep{naeini.m.2015} is used to quantify how well a model is calibrated. 
It computes the difference between the accuracy and confidence of the model, where accuracy is averaged across examples falling into confidence bins. 
A well-calibrated model will have a low ECE, as it will have a smaller difference between the predicted rate of success (the average confidence) and the actual rate of success (the average accuracy) of a given set of predictions. 
While ECE is a standard metric, it suffers from sensitivity to the number of confidence bins used \citep{ding.y.2020}.
To mitigate this, we use \citet{stengel-eskin-van-durme-2023-calibrated}'s implementation of \citet{ding.y.2020}'s adaptive binning approach, which dynamically adjusts bin sizes to reduce bias in the confidence estimate.

\paragraph{Area Under the Receiver Operator Characteristic curve (AUROC)} assesses the ability of the estimated confidence to distinguish correct and incorrect samples. 
AUROC measures the area under the curve formed by comparing the true positive rate to the false positive rate. 
If a model is well-calibrated, then there is some threshold for which we can separate predictions into correct predictions (above the threshold) and incorrect ones (below the threshold). 
In general, as we adjust the threshold there will be a tradeoff between true positives and false positives (e.g., a low threshold will result in a large number of false positives, while a high threshold will reduce the number of true positives). 
A higher AUROC score is better, with a perfect classifier achieving an AUROC of 1 while a random estimator would score 0.5. 
    
Figure~\ref{fig:calibration} illustrates Pearson correlations between two standard calibration metrics -- ECE and AUROC -- with \method{} performance. For Bash, 
we find no significant correlation with ECE and a moderate negative correlation with AUROC. 
For Webshop, neither metric is significantly correlated. 
Therefore, we conclude that a well-calibrated model is not a prerequisite for \method{}. 
This may be because calibration metrics do not measure ranking performance, which is central to our approach.
\label{append:calibration}

\section{Prompts}
\label{append:prompts}
We provide the prompts along with in-context examples supplied to the LLM for sampling trajectories for Bash and WebShop in \cref{prmpt:bash}, \cref{prmpt:web_select}, and \cref{prmpt:web_buy}. As mentioned in \cref{app:alf}, for ALFWorld, we use the prompts and in-context examples provided in \citet{yao2023react}.

\begin{figure*}
\centering
\begin{minipage}{0.95\textwidth}
\begin{lstlisting}[title=\texttt{Bash},basicstyle=\ttfamily\scriptsize,backgroundcolor=\color{prmpt}]
System: You are a helpful assistant expert specializing in BASH.
User: ## TASK DESCRIPTION
You are a BASH code generator helping me answer a question using BASH. 
I will ask you a question, and your task is to interact with a Bourne Shell system using BASH commands 
to come up with the answer. 

## RESPONSE FORMAT
Your response should be a BASH command. Format your BASH command as follows:
```BASH
Your BASH code here
```

DO NOT WRITE ANYTHING EXCEPT FOR CODE in your response.
Try ```sql
SHOW TABLES``` or ```sql
DESCRIBE <table_name> to learn more about the database```.

## OUTPUT DESCRIPTION
Given your BASH command input, the system will then give back output formatted as follows:

Output: <string>
Reward: [0, 1]

The output is the standard output from executing your BASH command.
The reward is a decimal value between 0 and 1, which tells you how close your BASH command is to the 
correct answer. 
The closer the reward is to 1, the closer your BASH command is to the correct answer.

You have to try to maximize the reward.

Query: "{query}".
Do not generate any output or reward.
Assistant: {Model Completion}
\end{lstlisting}
\end{minipage}
\caption{Prompt for Bash tasks.}
\label{prmpt:bash}
\end{figure*}

\begin{figure*}
\centering
\begin{minipage}{0.95\textwidth}
\begin{lstlisting}[title=\texttt{WebShop (adding a product to cart or selection)},basicstyle=\ttfamily\scriptsize,backgroundcolor=\color{prmpt}]
Instruction: Your task is to select a product that matches the user criteria and add it to your 'cart'.
You can use the following commands with the environment:
- search: To search for a query provided you are on a search page with a [search] button
- click: To click on a link enclosed within '[]'
- think: To add comments or thoughts without affecting the state of the webpage
- cart: To add an item to the cart from which it can be bought later.


Here are a few examples:
Criteria: I want to buy a 3 ounce bottle of citrus deodorant for sensitive skin and
clinically tested that is priced lower than $30 and natural.

Action: search[3 ounce bright citrus deodorant sensitive skin]
Observation: 
[Back to Search] 
Page 1 (Total results: 50) 
[Next >] 
[B078GWRC1J] 
Bright Citrus Deodorant by Earth Mama | Natural and Safe for Sensitive Skin, Pregnancy and
Breastfeeding, Contains Organic Calendula 3-Ounce 
$10.99 
[B078GTKVXY] 
Ginger Fresh Deodorant by Earth Mama | Natural and Safe for Sensitive Skin, Pregnancy and
Breastfeeding, Contains Organic Calendula 3-Ounce 
$10.99 
[B08KBVJ4XN] 
Barrel and Oak - Aluminum-Free Deodorant, Deodorant for Men, Essential Oil-Based Scent, 24
Hour Odor Protection, Cedar & Patchouli Blend, Gentle on Sensitive Skin (Mountain Sage, 2.7
oz, 2-Pack) 
$15.95  

Action: think[B078GWRC1J and B078GTKVXY are bright citrus deodorant less then 50 dollars. I
can check B078GWRC1J first to check if it matches all my requirements.]
Observation: OK.

Action: click[B078GWRC1J]
Observation: 
[Back to Search] 
[< Prev] 
scent [assorted scents][bright citrus][calming lavender][ginger fresh][simply non-scents]
size [travel set (4-pack)][3 ounce (pack of 1)][3-ounce (2-pack)]
Bright Citrus Deodorant by Earth Mama | Natural and Safe for Sensitive Skin, Pregnancy and
Breastfeeding, Contains Organic Calendula 3-Ounce 
Price: $10.99 
Rating: N.A. 
[Description] 
[Features] 
[Reviews] 
[Attributes]
[Buy Now]  

Action: think[I should explore the features of this product to verify it matches all my
requirements.]
Observation: OK.

Action: click[Features]
Observation:
[Back to Search]
[< Prev]
 NEW from Earth Mama (formerly Earth Mama Angel Baby), formulated especially for pregnancy,
 breastfeeding and sensitive skin  
 Contains organic grapefruit, tangerine and calendula  
 NO propylene glycol, artificial fragrance, parabens or aluminum  
 Dermatologist tested and clinically tested for irritation 
 Better than natural organic! NSF/ANSI 305 Certified by Oregon Tilth   

Action: think[B078GWRC1J matches my requirements: natural, citrus deodrant, for sensitive
skin, clinincally tested, and price less than $30. Thus, I should put it in my cart.]
Observation: OK.

Action: cart[B078GWRC1J]
Observation: OK.

Criteria: {}
\end{lstlisting}
\end{minipage}
\caption{Prompt for selecting and adding products to the cart in WebShop.}
\label{prmpt:web_select}
\end{figure*}
\begin{figure*}
\centering
\begin{minipage}{0.95\textwidth}
\begin{lstlisting}[title=\texttt{WebShop (buying a product)},basicstyle=\ttfamily\scriptsize,backgroundcolor=\color{prmpt}]
Instruction: Buy product [B078GWRC1J] that matches criteria: 3 ounce bottle of bright
citrus deodorant for sensitive skin, and price lower than 50.00 dollars
You can use the following commands with the environment:
- search: To search for a query provided you are on a search page with a [search] button
- click: To click on a link enclosed within '[]'. E.g., to buy use 'click[Buy Now]'
- think: To add comments or thoughts without affecting the state of the webpage

Action: think['I should be on the search page where I can see a list of products, I should
verify that.']
Observation: OK.

Action: load
Observation:
[Back to Search] 
Page 1 (Total results: 50) 
[Next >] 
[B078GWRC1J] 
Bright Citrus Deodorant by Earth Mama | Safe for Sensitive Skin, Pregnancy and
Breastfeeding, Contains Organic Calendula 3-Ounce 
$10.99 
[B08KBVJ4XN] 
Barrel and Oak - Aluminum-Free Deodorant, Deodorant for Men, Essential Oil-Based Scent, 24
Hour Odor Protection, Cedar & Patchouli Blend, Gentle on Sensitive Skin (Mountain Sage, 2.7 oz, 2-Pack) 
$35.95 
[B078GTKVXY] 
Ginger Fresh Deodorant by Earth Mama | Natural and Safe for Sensitive Skin, Pregnancy and
Breastfeeding, Contains Organic Calendula 3-Ounce 
$10.99 

Action: think['I can spot product [B078GWRC1J] and it is a clickable link since its
enclosed in square brackets. I should click on it.]
Observation: OK.

Action: click[B078GWRC1J]
Observation:
[Back to Search] 
[< Prev] 
scent [assorted scents][bright citrus][calming lavender][ginger fresh][simply non-scents]
size [travel set (4-pack)][3 ounce (pack of 1)][3-ounce (2-pack)]
Bright Citrus Deodorant by Earth Mama | Natural and Safe for Sensitive Skin, Pregnancy and
Breastfeeding, Contains Organic Calendula 3-Ounce 
Price: $10.99 
Rating: N.A. 
[Description] 
[Features] 
[Reviews] 
[Buy Now]

Action: think[I'm already on the product page.  I see multiple possible options enclosed in
'[]' for scent and size for this product. To match the input criteria, I should click on
options '[bright citrus]' for scent and '[3 ounce (pack of 1)]' for size one by one and
then buy in the end.]
Observation: OK.

Action: click[bright citrus]
Observation: You have clicked bright citrus. 

Action: click[3 ounce (pack of 1)]
Observation: You have clicked 3 ounce (pack of 1). 

Action: think[My task is to buy the product, for it should to click 'buy now']
Observation: OK.

Action: click[Buy Now]
Observation: You have clicked buy now.

Action: think[I finished buying the product. Task completed!]


Here is another task in which you need to buy a product. When you finish buying the product
with the most relevant choices, use 'think[Task completed']. If you cannot find the
matching options or proceed, think['Task failed']. Note that you can only click on text
enclosed in '[]' on the webpage. Everything else is only a description, not valid with  "click" action.

Instruction: Buy product [{}] that matches the criteria: {}
\end{lstlisting}
\end{minipage}
\caption{Prompt for buying products in WebShop.}
\label{prmpt:web_buy}
\end{figure*}

\end{document}